\documentclass{article}
\usepackage{spconf,amsmath,graphicx}
\usepackage[utf8]{inputenc}
\usepackage[T1]{fontenc}
\usepackage[english]{babel}
\usepackage{textcomp}
\usepackage{amsthm}
\usepackage{amsmath}
\usepackage{amssymb}
\usepackage{times}
\usepackage{soul}
\usepackage{epsfig}
\usepackage{amssymb}
\usepackage{multirow}
\usepackage{url,subfigure,graphicx,color,xcolor,booktabs,colortbl,threeparttable}
\usepackage{algorithm,algorithmic}
\usepackage{pifont}
\usepackage{enumitem}
\usepackage{array}
\usepackage{bm}
\usepackage{url}
\usepackage{ifpdf}
\usepackage{ifxetex}
\usepackage{caption}
\usepackage{hyperref}
\usepackage{cite}

\title{LongShortNet: Exploring Temporal and Semantic Features Fusion in Streaming Perception}

\vspace{-0.15in}
\name{\begin{tabular}{c}
Chenyang Li$^1$\sthanks{Equal contribution of C. Li, Z. Cheng, and J. He (in no particular order)},
Zhi-Qi Cheng$^2\footnotemark[1]$,
Jun-Yan He$^1\footnotemark[1]$, 
Pengyu Li$^1$,
\\
\textit{Bin Luo}$^1$\sthanks{Corresponding author}, 
\textit{Hanyuan Chen}$^1$, 
\textit{Yifeng Geng}$^1$,
\textit{Jin-Peng Lan}$^1$,
\textit{Xuansong Xie}$^1$
\vspace{-0.1in}
\end{tabular}}
\address{
$^1$DAMO Academy, Alibaba Group~~~~~~~
$^2$Carnegie Mellon University ~~~~~~~
\vspace{-0.15in}
}

\begin{document}
%
\maketitle
\begin{abstract}
Streaming perception is a fundamental task in autonomous driving that requires a careful balance between the latency and accuracy of the autopilot system. However, current methods for streaming perception are limited as they rely only on the current and adjacent two frames to learn movement patterns, which restricts their ability to model complex scenes, often leading to poor detection results. To address this limitation, we propose \textbf{LongShortNet}, a novel dual-path network that captures long-term temporal motion and integrates it with short-term spatial semantics for real-time perception. Our proposed LongShortNet is notable as it is the first work to extend long-term temporal modeling to streaming perception, enabling spatiotemporal feature fusion. We evaluate LongShortNet on the challenging Argoverse-HD dataset and demonstrate that it outperforms existing state-of-the-art methods with almost no additional computational cost.\footnote{Code is at https://github.com/LiChenyang-Github/LongShortNet.}
\end{abstract}
\vspace{-1mm}
\section{Introduction}
\vspace{-1mm}
\label{sec:introduction}
Autonomous driving requires the real-time perception of streaming video to react to motion changes, such as overtaking and turning. Different from traditional Video Object Detection (VOD) methods that focus on detecting and tracking objects in video frames~\cite{Ye_MultiFocus_ICASSP22,Chen_STFA_ICASSP20,Han_SeqNMS_CoRR16,Zhu_Deepflow_CVPR17,Zhu_Flowguide_ICCV17,Zhu_HighPerformance_CVPR18,Wang_PTSEFormer_ECCV22,Sun_LSPN_ECCV22,Chen_MEGA_CVPR2020,Deng_RDN_ICCV19,Han_HVRNet_ECCV20,Wu_CHP_ICCV19,Cui_TFB_ICCV21,Shvets_LRTR_ICCV19,Cheng_MM22}, Li et al. proposed a new autopilot perception task called streaming perception~\cite{LiWR_Streaming_ECCV20}. Streaming perception is a valuable tool for simulating realistic autonomous driving scenarios, offering the new metric of streaming Average Precision (sAP) to consistently evaluate accuracy and latency \cite{LiWR_Streaming_ECCV20}. Unlike offline VOD, streaming perception allows for real-time perception, opening up new possibilities for autonomous driving.

\begin{figure}[!ht]
  \centering
  \centerline{\includegraphics[width=0.80\linewidth]{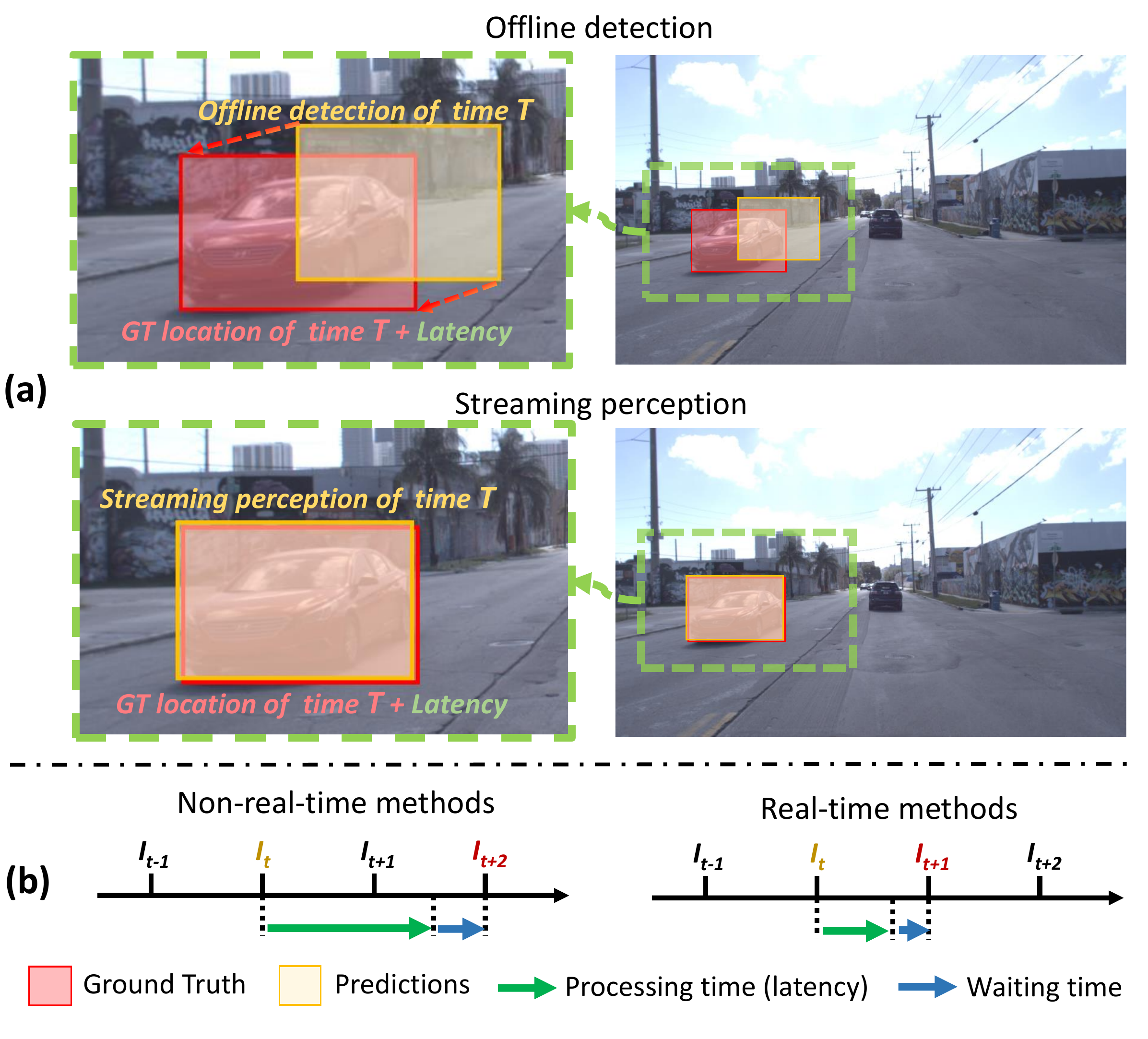}}
  \vspace{-4mm}
  \caption{\footnotesize {(a) Comparison between offline detection (VOD) and streaming perception, where the latter is real-time and can respond promptly to motion changes. (b) Timeline showing processing time.}}
\vspace{-7mm}
\label{fig:intro_sp}
\end{figure}

To enhance comprehension, Fig.~\ref{fig:intro_sp} provides a visual comparison between video-on-demand (VOD) and streaming perception, using colored bounding boxes to contrast real-world scenarios. Previous research \cite{Yang_Streamyolo_CVPR22} indicates that VOD methods, such as those presented in \cite{Chin_AdaScale_MLSys,He_QueryProp_AAAI22,He_TranVOD_MM21,SunHHR21_MAMBA_AAAI,Tang_HQlinking_TPAMI2020}, are vulnerable to errors due to offline detection delays. Although some VOD approaches have attempted to balance speed and accuracy by utilizing offline methods \cite{Ghosh_Adaptive_CoRR21}, streaming perception approaches, such as the one proposed in \cite{LiWR_Streaming_ECCV20}, only consider the last two frames to minimize latency, disregarding long-term temporal motions. Nonetheless, these methods have limitations in managing complicated motion and scene shifts due to their incapacity to account for the short-term spatial and long-term temporal aspects of video streams.

\begin{figure}[!t]
  \centering
  \centerline{\includegraphics[width=0.7\linewidth]{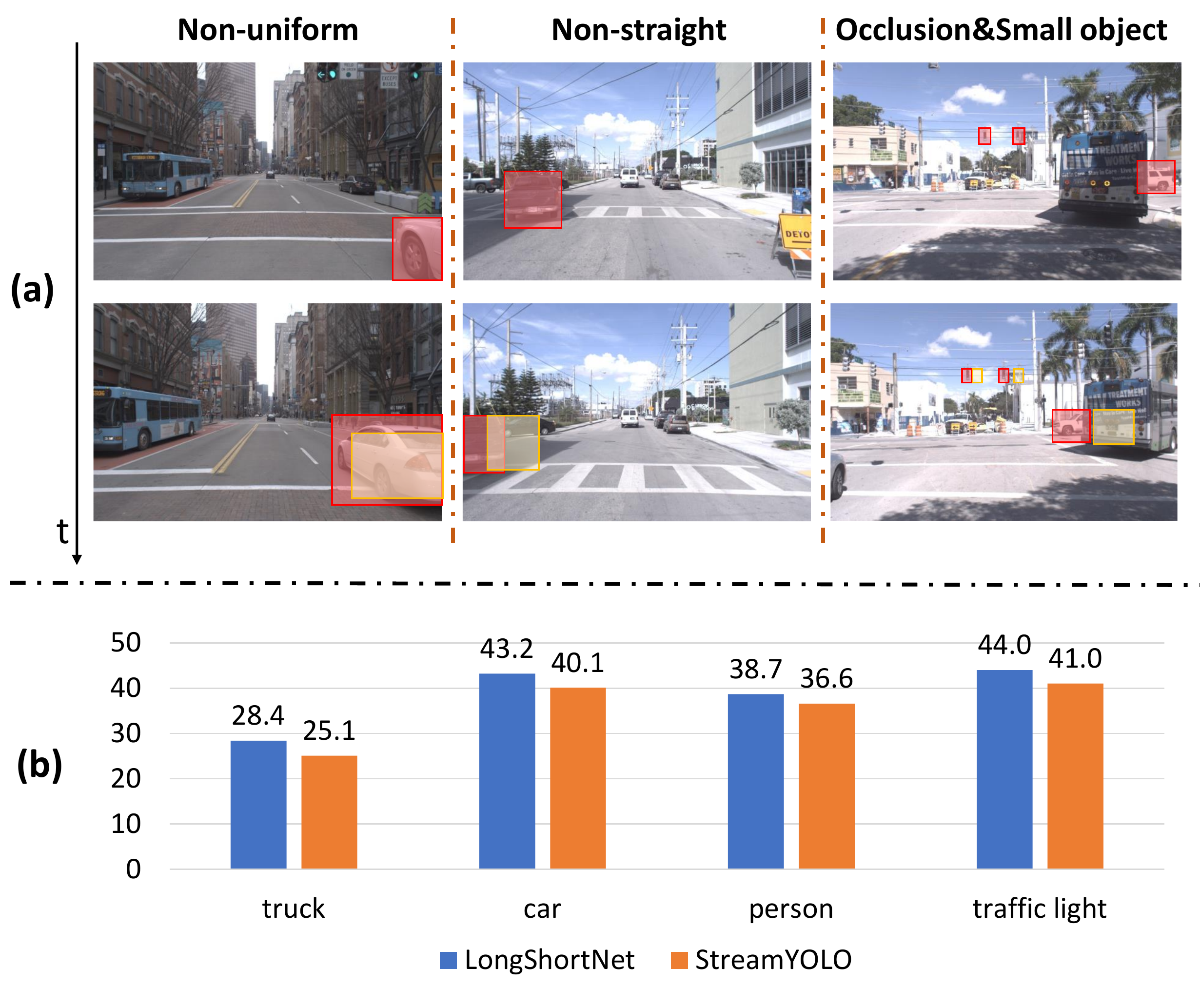}}
  \vspace{-4mm}
  \caption{\footnotesize (a) shows that StreamYOLO has suboptimal results, with red and orange boxes indicating ground truth and predictions. The comparison of sAP between LongShortNet and StreamYOLO is presented in (b). More examples can be found at {https://rebrand.ly/wgtcloo}.}
  \vspace{-5mm}
  \label{fig:complex_cases}
\end{figure}

Going further, we illustrate significant perceptual challenges arising from the lack of \textit{spatial semantics} and \textit{temporal motion}. Fig.~\ref{fig:complex_cases} displays a video stream captured by a car's front-view camera. The detected object's state in the video stream is influenced by its own motion and camera movement. Besides ideal uniform linear motion, actual video streams involve various challenges such as \textit{1)~non-uniform motion} (e.g., vehicle accelerating to overtake), \textit{2)~non-straight motion} (e.g., object and camera turning), \textit{3)~scene occlusion} (e.g., billboard and oncoming car occlusion), and \textit{4)~small objects} (e.g., cars and road signs in the distance). Undoubtedly, the motions and scenarios encountered in real autonomous driving are exceedingly complex and uncertain.

Faced with these concerns, StreamYOLO~\cite{Yang_Streamyolo_CVPR22} disregards the semantics and motion in video streams and only uses the last two frames as input (i.e., the current and previous frames). Due to the lack of spatial semantic and temporal motion cues, StreamYOLO is unable to handle complex scenes involving non-uniform, non-straight, and occluded objects. As shown in Fig.\ref{fig:complex_cases}(a), StreamYOLO\cite{Yang_Streamyolo_CVPR22} inaccurately locates objects in complex scenes and even misses occluded and small objects. The quantitative results presented in Fig.~\ref{fig:complex_cases}(b) demonstrate that incorporating temporal motion with spatial details can significantly improve perception accuracy, crucial for decision-making in autonomous driving, especially with increasing vehicle speed and scene complexity.

However, fusing spatiotemporal features while maintaining low latency is a challenging and open problem in streaming perception. Although prior techniques have investigated spatial-temporal aggregation~\cite{Wang_Motion_ECCV18,Zhu_Deepflow_CVPR17,Zhu_Flowguide_ICCV17,Zhu_HighPerformance_CVPR18, Xiao_STAligned_ECCV18,Bertasius_Sampling_ECCV18,He_NC2021,He_TIP21,jiang2012fast,Qiao_M22,huang2019decoupling,Lan_ICASSP2023} and spatial-temporal alignment~\cite{SunHHR21_MAMBA_AAAI,Tang_HQlinking_TPAMI2020,Cheng_ICCV19,Cheng_CVPR22} to extract enhanced temporal motion and spatial semantic information from support images/frames, these approaches are primarily tailored for offline settings and are not well-suited for online streaming perception. To address this issue, we propose a novel dual-path network named LongShortNet that calibrates long-term temporal with short-term spatial and models accurate motion coherence for streaming perception. Specifically, we investigate various fusion module setups and successfully introduce spatiotemporal fusion strategies from traditional VOD to streaming perception. With the support of spatiotemporal fusion and dilation-compression acceleration strategies, LongShortNet achieves a satisfactory trade-off between effectiveness and efficiency. In summary, our contributions can be summarized in threefold:
\vspace{-2mm}
\begin{itemize}[leftmargin=*]
\item To the best of our knowledge, LongShortNet is the first end-to-end model that learns long-term motion consistency for streaming perception, enabling the network to handle more complex autonomous driving scenarios.
\vspace{-1mm}
\item We propose a simple yet effective fusion module called Long-Short Fusion Module (LSFM), which fuses long-term temporal information with short-term spatial information using a dilation strategy, without requiring explicit supervision to learn motion consistency.
\vspace{-1mm}
\item LongShortNet achieves 37.1\% (normal size (600, 960)) and 42.7\% (large size (1200, 1920)) sAP without using any extra data, outperforming the existing state-of-the-art StreamYOLO \cite{Yang_Streamyolo_CVPR22} with almost the same time cost ({20.23 ms vs. 20.12 ms}). Moreover, LongShortNet works well on stricter metrics and small objects.
\vspace{-1mm}
\end{itemize}

\begin{figure*}[t]
	\begin{center}
		\includegraphics[width=0.8\linewidth]{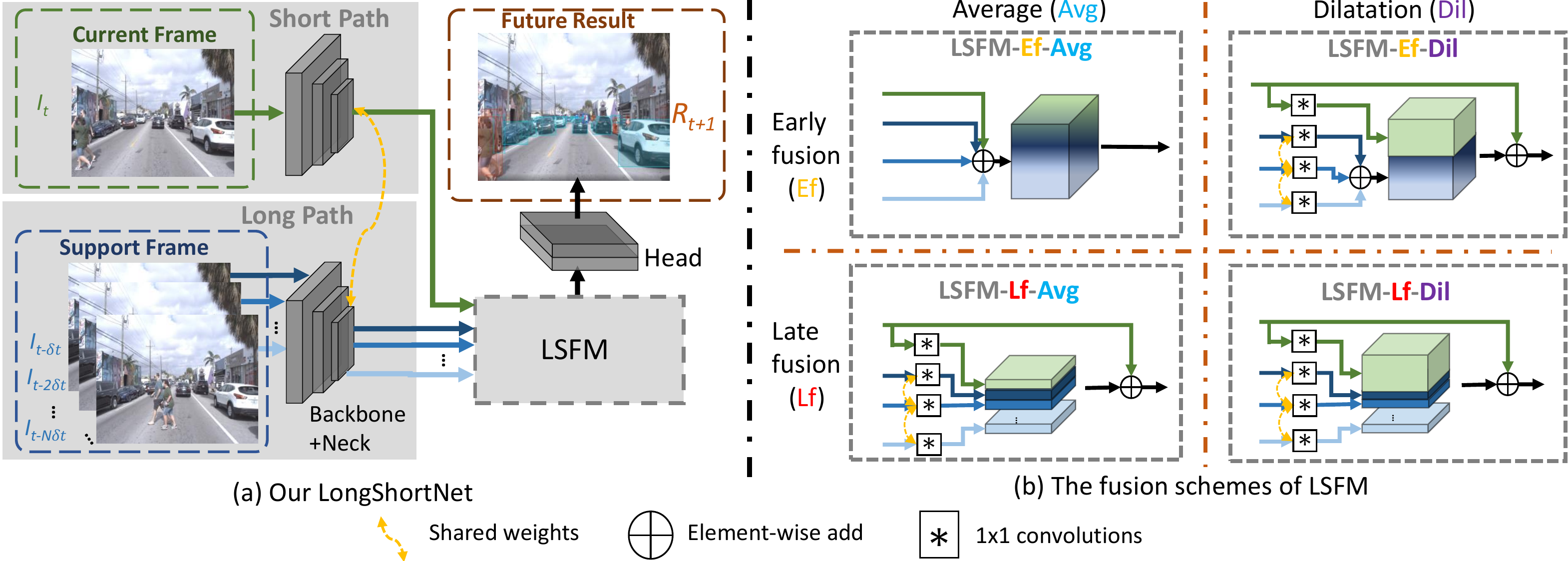}\\
	\end{center}
	\vspace{-0.25in}
	\caption{\footnotesize Illustration of our LongShortNet. (a) is an overview of LongShortNet. (b) shows the details of different fusion schemes of LSFM.}
	\vspace{-0.15in}
	\label{fig:framework}
\end{figure*}

\vspace{-4mm}
\section{Methodology}
\label{sec:method}
\vspace{-2mm}
\subsection{LongShortNet}
\vspace{-1mm}
\label{ssec:method_dpa}
We argue that \textit{spatial semantics} and \textit{temporal motion} are crucial for detecting complex movements such as non-uniform, non-linear, and occlusion. To this end, we propose \textit{LongShortNet}, which coherently models long-term temporal and fuses it with short-term semantics. LongShortNet consists of \textit{ShortPath} and \textit{LongPath}, as shown in Fig.~\ref{fig:framework}(a). The frame sequence captured by the camera is divided into the current frame and the support frame, which are fed into ShortPath and LongPath, respectively, to generate spatial and temporal features. The Long-Short Fusion Module \textit{(LSFM)} aggregates short-term spatial and long-term temporal information to capture motion consistency for representation learning. Finally, a detection \textit{head} predicts upcoming results based on the features produced by the LSFM.

Formally, ShortPath takes the current frame {\small$I_{t}$} as input and outputs spatial features {\small$F_{t}=\mathcal{F}(I_{t})$}, where {\small$\mathcal{F}(\cdot)$} is CNN networks, which includes the backbone (CSPDarknet-53~\cite{Alexey_YOLOV4_CoRR20}) and the neck~(PANet~\cite{Liu_PANet_CVPR18}).~Similarly, LongPath stores temporal features~{\small$F_{t-i\delta t}=\mathcal{F}(I_{t-i\delta t})$}, {\small$i \in [1, N]$} where {\small$N$} and {\small$\delta t$} denote the number of frames and time steps, respectively. {\small$\mathcal{F}(\cdot)$} represents the network of LongPath.
Note that the backbone of Short/Long paths is weight-shared.
By introducing tunable parameters {\small$N$} and {\small$\delta t$}, LongPath can capture more long-term temporal for fine movement reasoning. Then LSFM aggregates all features through {\small$F_{fuse}=\text{LSFM}(F_{t}, \dots, F_{t-N\delta t})$}, where {\small$F_{fuse}$} denotes the fused features generated by LSFM.
The details of {\small$\text{LSFM}(\cdot)$} are described in the next section. Finally, the results are acquired by {\small$D_{res}=\mathcal{H}(F_{fuse})$}, where {\small$\mathcal{H}$} denotes the detection head (TALHead~\cite{Yang_Streamyolo_CVPR22}) and {\small$D_{res}$} are predicted locations, scores, and categories.

\vspace{-2mm}
\subsection{Long Short Fusion Module}
\vspace{-1mm}
\label{ssec:method_lsfm}
The previous streaming perception work~\cite{Yang_Streamyolo_CVPR22} only \textit{roughly concatenates the features of the last two frames}, without exploiting temporal motion and spatial semantics.
We investigate a variety of feature aggregation ways, including 1) early fusion vs. late fusion and 2) average (equal weights) vs. dilatation (different weights).
In summary, we verified four types of LSFM as shown in Fig.~\ref{fig:framework}(b), denoted as LSFM-Ef-Avg, LSFM-Ef-Dil, LSFM-Lf-Avg, and LSFM-Lf-Dil.

\noindent {\textbf{Average-Early-Fusion.}}~The LSFM-Ef-Avg process fuses the spatial semantics of each frame in LSFM and outputs pre-averaged synthetic spatiotemporal features for the detection head.
This vanilla version allocates equal importance to the features of all frames, which is defined as,
\vspace{-2mm}
\begin{equation}\small
F_{fuse}=\sum_{i=1}^{N}F_{t-i\delta t}+F_{t},
\vspace{-2mm}
\label{eq:LSFM-Ef-Avg}
\end{equation}
where it counts all the features to fuse the current/historical spatial information directly and equally.

\noindent {\textbf{Dilatation-Early-Fusion}}.~For LSFM-Ef-Dil, we investigate different weighting schemes for feature fusion as,
\vspace{-2mm}
\begin{equation}\small
F_{fuse}=\text{Concat}(\mathcal{G}_{short}(F_{t}), \sum_{i=1}^{N} \mathcal{G}_{long}(F_{t-i\delta t}))+F_{t},
\vspace{-2mm}
\label{eq:LSFM-Ef-Dil}
\end{equation}
where {\small$\mathcal{G}$} denotes the {\small$1\times1$} convolution operation and {\small$\text{Concat}$} means the channel-wise concatenation.
Supposed that the channel dimensionality of {\small$F_{t}$} and {\small$F_{t-i\delta t}$} is {\small$d$}, all long-term temporal features are fused by addition before concatenating with the short-term spatial features.
In this case, the output channels numbers of {\small$\mathcal{G}_{short}(\cdot)$} and {\small$\mathcal{G}_{long}(\cdot)$} are both {\small$\lfloor d / 2 \rfloor$}.
Note that we also adopt a residual connection to add current spatial features to enhance the historical temporal features.

\noindent {\textbf{Average-Late-Fusion}}.~Contrary to the early fusion, LSFM-Lf-Avg fusion preserves the spatial semantic features of each frame separately and relies on the detection head to extract more high-level coherent features.
It instantly concatenates all features without discriminating between ShortPath and LongPath, which is defined as,
\vspace{-2mm}
\begin{equation}\small
F_{fuse}=\text{Concat}(\mathcal{G}_{avg}(F_{t}), \dots, \mathcal{G}_{avg}(F_{t-N\delta t}))+F_{t},
\vspace{-2mm}
\label{eq:LSFM-Lf-Avg}
\end{equation}
where the output channels number of {\small$\mathcal{G}_{avg}(\cdot)$} is {\small$\lfloor d / (1+N) \rfloor$}.
LSFM-Lf-Avg treats all features equally.

\noindent \textbf{{\textbf{Dilatation-Late-Fusion}}}.~We further propose LSFM-Lf-Dil, which enlarges the number of channels of ShortPath and forces LongShortNet to pay more attention to the current spatial information.
Specifically, LSFM-Lf-Dil is defined as,
\vspace{-1mm}
\begin{equation}\small
F_{fuse}=\text{Concat}(\mathcal{G}_{short}(F_{t}), \dots, \mathcal{G}_{long}(F_{t-N\delta t}))+F_{t},
\vspace{-1mm}
\label{eq:LSFM-Lf-Dil}
\end{equation}
where two {\small$1\times1$} convolution operations are employed to project {\small$F_{t}$} and {\small$F_{t-i\delta t}$} separately.~The output channels numbers of {\small$\mathcal{G}_{short}(\cdot)$} and {\small$\mathcal{G}_{long}(\cdot)$} are {\small$\lfloor d / 2 \rfloor$} and {\small$\lfloor d / 2N \rfloor$}.
After extensive experimental comparison, we finally chose Dilatation-Late-Fusion as LSFM and set {\small $N$} and {\small $\delta t$} to 3 and 1.
Please refer to Sec.~\ref{sec:ablation} for more details.

\begin{table}
\scriptsize
\centering
\caption{\footnotesize Comparison with non-real-time and real-time SOTA methods. `S=600/900' means the short edge of the input image is 600/900. `$^{\dagger}$' denotes adopting a larger input size (1200, 1920). Results shown in \textcolor[RGB]{0,102,255}{blue} font highlight the comparisons in stricter and harder metrics (sAP$_{75}$ and sAP$_{s}$).}
\vspace{-0.1in}
\setlength{\tabcolsep}{1.25mm}{
\begin{tabular}{c|c c c|c c c}
\hline
Methods & sAP & sAP$_{50}$ & sAP$_{75}$ & sAP$_{s}$ & sAP$_{m}$ & sAP$_{l}$ \\ \hline
\multicolumn{7}{c}{Non-real-time detector-based methods} \\ \hline
Streamer (S=900) \cite{LiWR_Streaming_ECCV20} & 18.2 & 35.3 & 16.8 & 4.7 & 14.4 & 34.6 \\
Streamer (S=600) \cite{LiWR_Streaming_ECCV20} & 20.4 & 35.6 & 20.8 & 3.6 & 18.0 & 47.2 \\ 
Streamer + AdaS \cite{Chin_AdaScale_MLSys,Ghosh_Adaptive_CoRR21} & 13.8 & 23.4 & 14.2 & 0.2 & 9.0 & 39.9 \\ 
Adaptive Streamer \cite{Ghosh_Adaptive_CoRR21}   & 21.3 & 37.3 & 21.1 & 4.4 & 18.7 & 47.1 \\ \hline
\multicolumn{7}{ c }{Real-time detector-based methods} \\ \hline
StreamYOLO-S \cite{Yang_Streamyolo_CVPR22} & 28.8 & 50.3 & 27.6 & 9.7 & 30.7 & 53.1 \\
StreamYOLO-M \cite{Yang_Streamyolo_CVPR22} & 32.9 & 54.0 & 32.5 & 12.4 & 34.8 & 58.1 \\ 
StreamYOLO-L \cite{Yang_Streamyolo_CVPR22} & 36.1 & 57.6 & \textcolor[RGB]{0,102,255}{35.6} & \textcolor[RGB]{0,102,255}{13.8} & 37.1 & 63.3 \\
\hline
LongShortNet-S     & 29.8 & 50.4 & 29.5 & 11.0 & 30.6 & 52.8 \\
LongShortNet-M     & 34.1 & 54.8 & 34.6 & 13.3 & 35.3 & 58.1 \\
LongShortNet-L     & 37.1 & 57.8 & \textbf{\textcolor[RGB]{0,102,255}{37.7}} & \textbf{\textcolor[RGB]{0,102,255}{15.2}} & 37.3 & \textbf{63.8} \\
LongShortNet-L$^{\dagger}$  & \textbf{42.7} & \textbf{65.4} & \textbf{45.0} & \textbf{23.9} & \textbf{44.8} & 61.7 \\ \hline
\end{tabular}}
\label{tab:comparing-sota}
\vspace{-6mm}
\end{table}

\begin{table*}
\scriptsize
\centering
\caption{{\footnotesize Details of LSFM (LSFM-Lf-Dil). The feature resolution is determined by the down-sampling rate, and the output channel numbers of {\scriptsize$\mathcal{G}{short}$} and {\scriptsize$\mathcal{G}{long}$} are computed as {\scriptsize$\lfloor d / 2 \rfloor$} and {\scriptsize$\lfloor d / 2N \rfloor$} in Sec.~\ref{ssec:method_lsfm}. Note that a 1x1 convolution is used to generate the outputs of LSFM if {\scriptsize$\lfloor d / 2 \rfloor + \lfloor d / 2N \rfloor \times N < d$}.}
}
\vspace{-3mm}
\begin{tabular}{c|c|c|cccc|cccc|cccc}
\hline
\multirow{2}{*}{Image size}    & \multirow{2}{*}{\begin{tabular}[c]{@{}c@{}c@{}} Down-\\sampling\\rate\end{tabular}} & \multirow{2}{*}{Resolution} & \multicolumn{4}{c|}{S-channels}                                                                                                                                                                                                                                                                                  & \multicolumn{4}{c|}{M-channels}                                                                                                                                                                                                                                                                                  & \multicolumn{4}{c}{L-channels}                                                                                                                                                                                                                                                                                   \\ \cline{4-15} 
                               &                                                                       &                             & \multicolumn{1}{c|}{\begin{tabular}[c]{@{}c@{}}LSFM\\ in\end{tabular}} & \multicolumn{1}{c|}{\begin{tabular}[c]{@{}c@{}}$\mathcal{G}_{short}$\\ out\end{tabular}} & \multicolumn{1}{c|}{\begin{tabular}[c]{@{}c@{}}$\mathcal{G}_{long}$\\ out\end{tabular}} & \begin{tabular}[c]{@{}c@{}}LSFM\\ out\end{tabular} & \multicolumn{1}{c|}{\begin{tabular}[c]{@{}c@{}}LSFM\\ in\end{tabular}} & \multicolumn{1}{c|}{\begin{tabular}[c]{@{}c@{}}$\mathcal{G}_{short}$\\ out\end{tabular}} & \multicolumn{1}{c|}{\begin{tabular}[c]{@{}c@{}}$\mathcal{G}_{long}$\\ out\end{tabular}} & \begin{tabular}[c]{@{}c@{}}LSFM\\ out\end{tabular} & \multicolumn{1}{c|}{\begin{tabular}[c]{@{}c@{}}LSFM\\ in\end{tabular}} & \multicolumn{1}{c|}{\begin{tabular}[c]{@{}c@{}}$\mathcal{G}_{short}$\\ out\end{tabular}} & \multicolumn{1}{c|}{\begin{tabular}[c]{@{}c@{}}$\mathcal{G}_{long}$\\ out\end{tabular}} & \begin{tabular}[c]{@{}c@{}}LSFM\\ out\end{tabular} \\ \hline
\multirow{3}{*}{(600,960,3)}   & /8                                                                    & (75,120)                    & \multicolumn{1}{c|}{128}                                               & \multicolumn{1}{c|}{64}                                                                  & \multicolumn{1}{c|}{21}                                                                 & 128                                                & \multicolumn{1}{c|}{192}                                               & \multicolumn{1}{c|}{96}                                                                  & \multicolumn{1}{c|}{32}                                                                 & 192                                                & \multicolumn{1}{c|}{256}                                               & \multicolumn{1}{c|}{128}                                                                 & \multicolumn{1}{c|}{42}                                                                 & 256                                                \\ \cline{2-15} 
                               & /16                                                                   & (38,60)                     & \multicolumn{1}{c|}{256}                                               & \multicolumn{1}{c|}{128}                                                                 & \multicolumn{1}{c|}{42}                                                                 & 256                                                & \multicolumn{1}{c|}{384}                                               & \multicolumn{1}{c|}{192}                                                                 & \multicolumn{1}{c|}{64}                                                                 & 384                                                & \multicolumn{1}{c|}{512}                                               & \multicolumn{1}{c|}{256}                                                                 & \multicolumn{1}{c|}{85}                                                                 & 512                                                \\ \cline{2-15} 
                               & /32                                                                   & (19,30)                     & \multicolumn{1}{c|}{512}                                               & \multicolumn{1}{c|}{256}                                                                 & \multicolumn{1}{c|}{85}                                                                 & 512                                                & \multicolumn{1}{c|}{768}                                               & \multicolumn{1}{c|}{384}                                                                 & \multicolumn{1}{c|}{128}                                                                & 768                                                & \multicolumn{1}{c|}{1024}                                              & \multicolumn{1}{c|}{512}                                                                 & \multicolumn{1}{c|}{170}                                                                & 1024                                               \\ \hline
\multirow{3}{*}{(1200,1920,3)} & /8                                                                    & (150,240)                           & \multicolumn{1}{c|}{128}                                               & \multicolumn{1}{c|}{64}                                                                  & \multicolumn{1}{c|}{21}                                                                 & 128                                                & \multicolumn{1}{c|}{192}                                               & \multicolumn{1}{c|}{96}                                                                  & \multicolumn{1}{c|}{32}                                                                 & 192                                                & \multicolumn{1}{c|}{256}                                               & \multicolumn{1}{c|}{128}                                                                 & \multicolumn{1}{c|}{42}                                                                 & 256                                                \\ \cline{2-15} 
                               & /16                                                                   & (75,120)                           & \multicolumn{1}{c|}{256}                                               & \multicolumn{1}{c|}{128}                                                                 & \multicolumn{1}{c|}{42}                                                                 & 256                                                & \multicolumn{1}{c|}{384}                                               & \multicolumn{1}{c|}{192}                                                                 & \multicolumn{1}{c|}{64}                                                                 & 384                                                & \multicolumn{1}{c|}{512}                                               & \multicolumn{1}{c|}{256}                                                                 & \multicolumn{1}{c|}{85}                                                                 & 512                                                \\ \cline{2-15} 
                               & /32                                                                   & (38,60)                           & \multicolumn{1}{c|}{512}                                               & \multicolumn{1}{c|}{256}                                                                 & \multicolumn{1}{c|}{85}                                                                 & 512                                                & \multicolumn{1}{c|}{768}                                               & \multicolumn{1}{c|}{384}                                                                 & \multicolumn{1}{c|}{128}                                                                & 768                                                & \multicolumn{1}{c|}{1024}                                              & \multicolumn{1}{c|}{512}                                                                 & \multicolumn{1}{c|}{170}                                                                & 1024  \\
                               \hline
\end{tabular}
\vspace{-6mm}
\label{tab:lsfm-details}
\end{table*}

\vspace{-2mm}
\section{Experiments}
\label{sec:exp}
\vspace{-2mm}
\subsection{Dataset and Metric}
\vspace{-2mm}
Our experiments are conducted on the Argoverse-HD dataset, which is a streaming perception dataset. Following previous works~\cite{LiWR_Streaming_ECCV20, Yang_Streamyolo_CVPR22}, we utilize streaming Average Precision (sAP) as the evaluation metric. We also use the same train/val split as previous work~\cite{LiWR_Streaming_ECCV20, Yang_Streamyolo_CVPR22}.

\vspace{-3mm}
\subsection{Implementation Details}
\vspace{-2mm}
To fine-tune LongShortNet, we utilize a COCO pre-trained model and train for 8 epochs using a batch size of 16. Our base detectors are YOLOX-S, YOLOX-M, and YOLOX-L, corresponding to LongShortNet-S, LongShortNet-M, and LongShortNet-L, respectively, and we adopt the same loss function as in~\cite{Yang_Streamyolo_CVPR22}. The resolution and channel numbers of the LSFM, determined jointly by model size and $\mathcal{G}(\cdot)$, are provided in Tab.\ref{tab:lsfm-details} for $N=3$. To ensure the real-time performance of the network, we make slight modifications to the buffer scheme proposed in\cite{Yang_Streamyolo_CVPR22}. The experiments are carried out on 4 NVIDIA V100 GPUs, with all other hyperparameters following the previous work~\cite{Yang_Streamyolo_CVPR22}.

\begin{table}
\scriptsize
\centering
\caption{\footnotesize {Inference time (NVIDIA V100) and FLOPs (input size (600, 960)) for LongShortNet vs. StreamYOLO. The increased values are shown in \textcolor[RGB]{0,102,255}{blue}.}}
\vspace{-0.1in}
\setlength{\tabcolsep}{1.35mm}{
\begin{tabular}{c|c c}
\hline
Methods & Inference time (ms/frame) & FLOPs (G) \\ \hline
StreamYOLO-L  \cite{Yang_Streamyolo_CVPR22}   & 20.12          & 222.52 \\ \hline
LongShortNet-L                             & 20.23 (\textcolor[RGB]{0,102,255}{+0.11})          & 223.11 (\textcolor[RGB]{0,102,255}{+0.59}) \\ \hline
StreamYOLO-M  \cite{Yang_Streamyolo_CVPR22}       & 18.15      & 105.72 \\ \hline
LongShortNet-M     & 18.31 (\textcolor[RGB]{0,102,255}{+0.16})      & 106.05 (\textcolor[RGB]{0,102,255}{+0.33}) \\ \hline
StreamYOLO-S  \cite{Yang_Streamyolo_CVPR22}       & 14.22      & 38.55 \\ \hline
LongShortNet-S     & 14.34 (\textcolor[RGB]{0,102,255}{+0.12})      & 38.69 (\textcolor[RGB]{0,102,255}{+0.14}) \\ \hline
\end{tabular}}
\vspace{-2mm}
\label{table:computation}
\end{table}

\begin{table}[!ht]
\scriptsize 
\centering
\caption{{\footnotesize The exploration of dilation channel ratio in LSFM-Lf-Dil.}}
\vspace{-0.1in}
\setlength{\tabcolsep}{1.35mm}{
\begin{tabular}{c|c c c}
\hline
Dilation channel ratio & LongShortNet-S & LongShortNet-M & LongShortNet-L \\ \hline
0.25                           & 28.9          & 33.7          & 36.7             \\ \hline
0.5                           & \textbf{29.8}          & \textbf{34.1}          & \textbf{37.1}                          \\ \hline
0.75                          & 28.9          & 33.5          & 36.0                          \\ \hline
\end{tabular}}
\vspace{-4mm}
\label{table:dilation-ratio}
\end{table}

\vspace{-3mm}
\subsection{Comparison with SOTA Methods}
\vspace{-2mm}
We compared LongShortNet with state-of-the-art streaming perception methods on the Argoverse-HD~\cite{LiWR_Streaming_ECCV20} dataset in Tab.~\ref{tab:comparing-sota}, which includes both non-real-time and real-time approaches. LongShortNet outperforms all real-time and non-real-time baselines with a performance of {\small$37.1\%$}. Interestingly, LongShortNet also achieved an improved sAP of {\small$42.7\%$} when using larger input sizes without using extra data.

Specifically, Streamer~\cite{LiWR_Streaming_ECCV20}, which is a non-real-time detector-based method that integrates association and dynamic scheduling to overcome temporal aliasing, only achieves {\small$20.4\%$} sAP. StreamYOLO~\cite{Yang_Streamyolo_CVPR22}, a recent end-to-end model that produces results directly in the inference phase, achieves {\small$36.1\%$} sAP. LongShortNet achieves similar performance to StreamYOLO but with an almost negligible increase in latency (around \textbf{0.1+ ms}), despite integrating triple the historical features of StreamYOLO.

It is also worth noting that LongShortNet outperforms other methods in stricter metrics (sAP${75}$) and for small objects (sAP${s}$), indicating that long-term temporal features improve performance in complex scenarios. However, sAP$_{l}$ decreases when adopting a large input size, which is due to insufficient receptive field size.

\vspace{-3mm}
\subsection{Ablation Study}
\vspace{-2mm}
\label{sec:ablation}
\noindent \textbf{Temporal range effects}.~We tested the impact of temporal range by altering parameters $N$ and $\delta t$, as displayed in Tab.~\ref{tab:temporal-range}. Here, $(0,-)$ denotes using only the current frame, while $(1, 1)$ corresponds to StreamYOLO~\cite{Yang_Streamyolo_CVPR22}. LongShortNet-S/M/L achieved optimal performance with $(N,\delta t)$ values of $(3,1)$, $(4,2)$, and $(3,1)$, suggesting that long-term temporal cues effectively model motion consistency.

\noindent \textbf{Dilation channel ratio effects}.~We evaluated the influence of the dilation channel ratio for LSFM-Lf-Dil with $N=3$ and $d=1$, as shown in Tab.~\ref{table:dilation-ratio}. The S/M/L models reached their peak performance with a ratio of 0.5, emphasizing the importance of short-term spatial information. Performance declined when the ratio was too large, leading us to select a ratio of 0.5 for the dilation operation.

\noindent \textbf{Aggregation setup effects}.~To identify suitable long-term fusion strategies, we conducted ablation studies on LSFM. Tab.~\ref{table:lsfm-fusion} reveals that the late fusion with dilation strategy yielded the most favorable results for the S/M/L models, indicating that the proposed LSFM-Lf-Dil module effectively aggregates short-term semantic and long-term temporal features. Moreover, removing the residual connection led to a performance drop, highlighting the importance of current semantic features.

\noindent \textbf{Efficiency analysis}.~As the number of input frames increases, so does the computational cost.~However, employing the buffer scheme~\cite{Yang_Streamyolo_CVPR22}, LongShortNet requires only three additional $1\times1$ convolutional operations compared to StreamYOLO~\cite{Yang_Streamyolo_CVPR22} to store feature maps generated by the neck.~Tab.~\ref{table:computation} presents the inference time and FLOPs, demonstrating that LongShortNet, with long-range motion consistency capturing, incurs virtually no additional time cost compared to StreamYOLO~\cite{Yang_Streamyolo_CVPR22}.

\begin{table}
\scriptsize 
\centering
\caption{\footnotesize The ablation study of $N$ and $\delta t$. The best two results, the worst one, and the improvements are shown in \textcolor[RGB]{0,153,51}{green}, \textcolor[RGB]{238,153,34}{orange}, \textcolor[RGB]{255,100,97}{red} and \textcolor[RGB]{0,102,255}{blue} fonts.}
\vspace{-0.1in}
\setlength{\tabcolsep}{1.35mm}{
\begin{tabular}{c|c c c}
\hline
($N$, $\delta t$) & LongShortNet-S & LongShortNet-M & LongShortNet-L \\ \hline
(0, -)                           & \textcolor[RGB]{255,100,97}{26.8}          & \textcolor[RGB]{255,100,97}{29.8}          & \textcolor[RGB]{255,100,97}{32.6}                          \\ \hline
(1, 1)                           & 28.7          & 33.5          & 36.1                          \\ \hline
(1, 2)                           & 28.9          & 33.8          & 36.2             \\ \hline
(2, 1)                           & 29.2          & 34.0          & 36.7             \\ \hline
(2, 2)                           & 29.1          & 34.0          & 36.4             \\ \hline
(3, 1)                           & \textbf{\textcolor[RGB]{0,153,51}{29.8}}   $\uparrow$ \textbf{\textcolor[RGB]{0,102,255}{+3.0}}       & \textcolor[RGB]{238,153,34}{34.1}          & \textbf{\textcolor[RGB]{0,153,51}{37.1}} $\uparrow$ \textbf{\textcolor[RGB]{0,102,255}{+4.5}}             \\ \hline
(3, 2)                           & \textcolor[RGB]{238,153,34}{29.7}          & \textcolor[RGB]{238,153,34}{34.1}          & 36.7             \\ \hline
(4, 1)                           & 29.2          & 34.0          & \textcolor[RGB]{238,153,34}{37.0}            \\ \hline
(4, 2)                           & 28.8          & \textbf{\textcolor[RGB]{0,153,51}{34.2}}  $\uparrow$ \textbf{\textcolor[RGB]{0,102,255}{+4.4}}        & 36.6                          \\ \hline
(5, 1)                           & 29.3          & 33.4          & 36.9             \\ \hline
(5, 2)                           & 28.1          & 33.6          & 36.3             \\ \hline
\end{tabular}}
\vspace{-2mm}
\label{tab:temporal-range}
\end{table}

\begin{table}
\scriptsize 
\centering
\caption{\footnotesize The ablation study of LSFM. `$^{\ast}$' denotes removing residual connection. The improvements are shown in \textcolor[RGB]{0,102,255}{blue} font.}
\vspace{-0.1in}
\setlength{\tabcolsep}{1.35mm}{
\begin{tabular}{c|c c c}
\hline
LSFM & LongShortNet-S & LongShortNet-M & LongShortNet-L \\ \hline
LSFM-Ef-Avg                           & 27.1          & 31.7          & 35.3             \\ \hline
LSFM-Ef-Dil                           & 28.6          & 33.2          & 36.1                          \\ \hline
LSFM-Lf-Avg                          & 28.9          & 33.7          & 36.7                          \\ \hline
LSFM-Lf-Dil                          & \textbf{29.8} (\textcolor[RGB]{0,102,255}{+2.7})         & \textbf{34.1} (\textcolor[RGB]{0,102,255}{+2.4})         & \textbf{37.1} (\textcolor[RGB]{0,102,255}{+1.8})                         \\ \hline
LSFM-Lf-Dil$^{\ast}$                 & 27.6          & 32.1          & 35.3                          \\ \hline
\end{tabular}}
\vspace{-4mm}
\label{table:lsfm-fusion}
\end{table}

\vspace{-3mm}
\section{Conclusion}
\vspace{-2mm}
\label{sec:conclusion}
We present LongShortNet, a real-time perception model skillfully fusing short-term spatial semantics and long-term temporal motion cues, achieving a superior balance between accuracy and efficiency. Our in-depth ablation studies underscore the vital roles of temporal range, dilation channel ratio, and aggregation setups in fusing spatiotemporal features, providing essential insights for future research. Surpassing state-of-the-art methods, LongShortNet excels in streaming Average Precision (sAP) on the Argoverse-HD dataset, with minimal computational overhead, emphasizing its potential for autonomous driving applications. In future work, we aim to integrate explicit motion consistency constraints based on geometric context to further refine the performance.
\vfill\pagebreak

{\small
\bibliographystyle{IEEEbib}
\bibliography{refs}}

\end{document}